\documentclass{article}

\PassOptionsToPackage{numbers, sort&compress}{natbib}
 \usepackage[preprint]{neurips_2026}

\usepackage[utf8]{inputenc} 
\usepackage[T1]{fontenc}    
\usepackage{xcolor}         
\definecolor{cvprblue}{rgb}{0.21,0.49,0.74}
\usepackage[pagebackref,breaklinks,colorlinks,allcolors=cvprblue]{hyperref}
\usepackage{url}            
\usepackage{booktabs}       
\usepackage{amsfonts}       
\usepackage{nicefrac}       
\usepackage{microtype}      
\usepackage{enumitem}
\usepackage[font=small]{caption}
\newcommand{\com}[1]{}

\renewcommand{\paragraph}[1]{\vspace{.4em}\noindent\textbf{#1.}}

\usepackage{xspace}
\newcommand{\paper}{H-Flow\xspace}
\newcommand{\dynact}{DynAct4D\xspace}

\usepackage{soul}
\setuldepth{foobar}

\usepackage{tikz}

\newcommand{\macaption}[1]{\small \textbf{#1}\xspace}
\newcommand{\micaption}[1]{\small #1}

\newcommand{\best}{\textbf{bold}}
\newcommand{\second}{\underline{\textit{underlined}}}

\usepackage{pifont}

\definecolor{carmine}{rgb}{0.59, 0.0, 0.09}
\definecolor{teal}{rgb}{0.0, 0.5, 0.5}
\definecolor{navy}{rgb}{0.129, 0.255, 0.451}
\definecolor{darkblue}{rgb}{0.129, 0.255, 0.451}
\definecolor{darkgreen}{rgb}{0.165, 0.373, 0.357}
\definecolor{darkpink}{rgb}{0.494, 0.122, 0.278}
\definecolor{mint}{rgb}{0.92, 0.96, 0.90}

\definecolor{appendixbeige}{RGB}{181, 146, 102}

\usepackage{nicefrac}

\usepackage{amsmath}
\usepackage{placeins}
\usepackage{makecell}
\usepackage{tabularx}
\usepackage{calc}
\usepackage[table]{xcolor}
\usepackage{subcaption}
\usepackage{multirow}

\usepackage[capitalise]{cleveref}
\crefname{section}{Sec.}{Secs.}
\crefname{subsection}{Sec.}{Secs.}
\crefname{subsubsection}{Sec.}{Secs.}
\crefname{table}{Tab.}{Tabs.}
\crefname{figure}{Fig.}{Figs.}
\crefname{equation}{Eq.}{Eqs.}

\let\oldcite\cite
\renewcommand{\cite}[1]{{\small\oldcite{#1}}}

\title{H-Flow: Self-supervised Human Scene Flow via Physics-inspired Joint Multi-modal Learning}

\author{Zhanbo Huang$^1$, Xiaoming Liu$^{1,2}$, Yu Kong$^{1}$ \\
{\normalsize $^1$Michigan State University \quad $^2$University of North Carolina, Chapel Hill} \\
{\tt\small $^1$\{huang247, yukong\}@msu.edu \quad $^2$liuxm@cs.unc.edu}}

\begin{document}

\maketitle

\begin{abstract}
Parametric human models capture global pose but cannot represent the non-rigid surface dynamics of clothing and soft tissue.
Generic scene flow estimates dense motion but breaks down on articulated bodies, where pixel-level supervision is also intractable to acquire.
We introduce \emph{\paper}, a dense human scene flow that captures both skeletal kinematics and surface deformation.
A unified multi-head transformer estimates flow from monocular video, jointly predicting pose and depth as companion outputs.
The challenge lies in the lack of supervision.
In place of unattainable labels, we anchor the network in the physics of human motion, encoding geometric, structural, and biomechanical priors as cross-modal training objectives.
We further introduce \emph{\dynact}, a high-fidelity synthetic benchmark providing dense flow annotations across diverse subjects, garments, and motions.
On standard benchmarks, \paper outperforms scene-flow and parametric baselines, and generalizes zero-shot to in-the-wild video.
Code, models, and the \dynact benchmark will be released upon publication.
\end{abstract}
\section{Introduction}
\label{sec:intro}

\begin{figure}[t]
    \centering
    \includegraphics[width=0.94\linewidth]{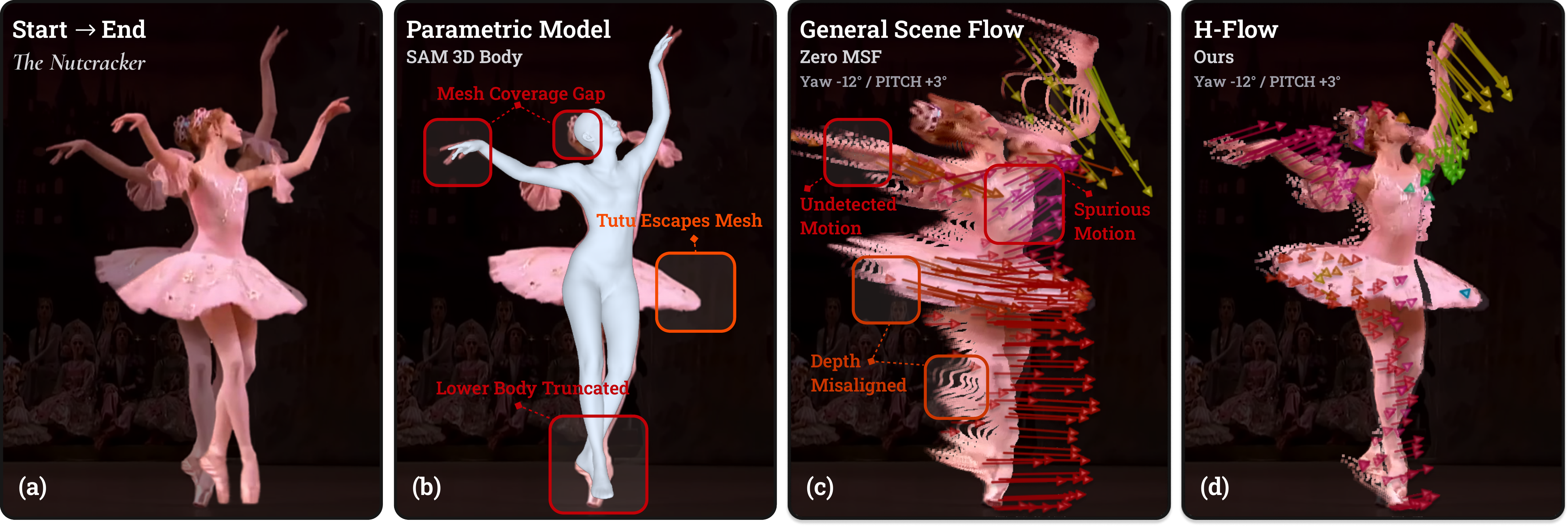}
    \caption{\macaption{Comparison of paradigms for human motion perception on an in-the-wild ballet performance.}
        \micaption{Evaluated zero-shot on a frame from \emph{The Nutcracker} {\footnotesize (Mariinsky Ballet)}.
        \emph{(a)} The input video frame.
        \emph{(b)} Parametric models (SAM 3D Body~\cite{yang2026sam3dbody}) recover global pose, but their mesh excludes the tutu and truncates extremities.
        \emph{(c)} Generic scene flow (Zero MSF~\cite{liang2025zeroshot}) yields undetected motion on the dancer, spurious vectors in static regions, and misaligned depth.
        \emph{(d)} Our \paper produces accurate dense 3D motion across body and garment, anchored in physical priors and learned without any flow ground truth.
        \emph{The textured backgrounds in (c) and (d) are 3D point clouds reconstructed from each method's predicted depth. See \cref{sec:app_teaser} for additional viewpoints.}}
    }
    \label{fig:teaser}
    \vspace{-1em}
\end{figure}

Capturing how a person moves involves more than tracking joints.
The body's surface deforms in ways skeletal kinematics alone cannot describe.
A complete account needs both skeletal structure and dense surface detail.
Parametric models~\cite{loper2015smpl,pavlakos2019expressive,xu2020ghum,osman2020star,osman2022supr,ferguson2025mhr} have become the dominant paradigm, with recent methods enabling robust recovery of 3D pose and shape from a single image~\cite{goel2023hmr2,yang2026sam3dbody}.
Yet their surfaces are bound to linear blend skinning, where each vertex is a weighted sum of bone transformations.
This cannot represent motion that deviates from the skeleton, such as the swirl of a skirt or the deformation of soft tissue~\cite{bertiche2021pbns,srivastava2026physkin}.
These details are essential for behavior understanding and realistic modeling, yet remain out of reach for skeleton-bound methods (\cref{fig:teaser}{\small b}).

Dense 3D scene flow~\cite{vedula1999three,liu2019flownet3d,puy2020flot} offers an alternative path.
It estimates a motion vector for every visible point, providing the density needed to represent clothed humans.
But applying it to articulated humans faces a fundamental structural barrier (\cref{fig:teaser}{\small c}).
Most methods rely on local rigid-body assumptions to regularize the ill-posed matching problem~\cite{lin2024icp,lin2025voteflow}.
Such rigidity priors break down on highly articulated, non-rigid human motion, smearing predictions at limb tips and garment edges.
Human-region accuracy thus consistently lags behind generic-scene performance.

To close this gap, we introduce \paper, a dense scene flow specialized for human bodies.
Every visible human point carries its own 3D motion vector, providing the per-point density needed to capture garment folds and soft-tissue deformation.
Beyond density, \paper targets the skeletal accuracy of parametric models, expressing articulated motion without resorting to global rigid-body assumptions that contradict anatomical reality.
Realizing such a representation requires acquiring human-aligned motion structure under no flow supervision.
Pose, depth, and flow appear as three separate modalities, yet they are different projections of one physical system: a human body moving through space.
Body geometry shapes the depth field and skeletal kinematics shape the surface flow, with both couplings arising from the body's physical structure rather than any algorithmic choice.
This view turns self-supervision from constraining a modality against itself into constraining modalities against the physics that binds them.
This goal lies beyond existing learning paradigms.
Supervised methods require dense human flow ground truth that is not available.
Self-supervised scene flow methods~\cite{lin2024icp,lin2025voteflow} rely on the rigid-body assumption we set out to discard.
We therefore design a new learning paradigm.
\paper, pose, and depth are jointly estimated in a single architecture, with geometric, structural, and biomechanical priors encoded as cross-modal training objectives.
The system trains in a fully self-supervised manner with no flow ground truth.
To inject absolute metric scale, we further introduce SAM 3D Body and Depth Pro~\cite{yang2026sam3dbody,bochkovskii2024depthpro} as anchors.
During training, flow progressively aligns with human kinematics.
Pose surpasses its anchor on absolute joint position, and depth significantly outperforms its anchor within the body region.
The underlying principle is cross-modal mutual benefit.
Pose acquires metric awareness through depth and flow, while depth concentrates capacity on the body region through pose and flow.

To quantitatively evaluate \paper as a dense human scene flow representation, no existing dataset suffices.
Scene flow benchmarks~\cite{geiger2012kitti,mayer2016flyingthings3d,butler2012naturalistic} target rigid or generic scenes rather than human bodies, and human motion datasets~\cite{ionescu2014human3,vonmarcard2018recovering,mahmood2019amass} lack dense flow annotations.
Acquiring human flow ground truth in 2D has been addressed by the optical flow community through synthetic benchmarks~\cite{butler2012naturalistic,mayer2016flyingthings3d}, yet the 3D case is harder, with real-world capture intractable on non-rigid clothed bodies.
Following this synthetic route, we build \dynact on the MetaHuman framework~\cite{epic2024metahuman}, generating photorealistic, anatomically diverse human characters with simulated cloth physics inside Unreal Engine~\cite{epic2022unreal}.
\dynact features characters across genders, ethnicities, and body types, dressed in garments of varying length and fit, performing a wide range of actions across multiple scenes and viewpoints.
Every frame provides per-pixel flow ground truth, offering a rigorous testbed for human scene flow.

On standard benchmarks, \paper outperforms scene-flow and parametric baselines, and \dynact provides rigorous evaluation of dense human motion.
The same model generalizes zero-shot to in-the-wild ballet and everyday dance videos.
In summary, our contributions are fourfold.
\begin{itemize}[leftmargin=0.5cm, itemsep=2pt, topsep=2pt, parsep=0pt]
    \item We introduce \paper, a dense scene flow specialized for human bodies that combines the skeletal accuracy of parametric models with the per-point density of generic scene flow.
    \item We propose a self-supervised learning paradigm that encodes geometric, structural, and biomechanical priors as cross-modal training objectives between pose, depth, and flow, training without flow labels and yielding mutual gains across all three modalities.
    \item We construct \dynact, a high-fidelity synthetic benchmark with dense flow ground truth across diverse subjects, garments, and actions.
    \item We demonstrate that \paper generalizes zero-shot to in-the-wild humans, indicating that physics-inspired cross-modal coupling captures transferable regularities of human motion rather than dataset-specific patterns.
\end{itemize}
\section{Related Work}
\label{sec:related}

\paragraph{Parametric Models and Scene Flow}
Parametric models like SMPL~\cite{loper2015smpl} and SMPL-X~\cite{pavlakos2019expressive} provide strong geometric priors for 3D human shape and pose, with deep estimation methods enabling robust recovery from monocular video~\cite{kanazawa2018hmr,goel2023hmr2,shin2024wham,cai2023smplerx,yang2026sam3dbody}.
However, their reliance on Linear Blend Skinning (LBS) cannot model surfaces that move independently of the skeletal topology, such as loose clothing.
Recent hybrid and implicit methods attempt to bypass this limitation, either by augmenting parametric meshes with learned cloth layers~\cite{tan2025dressrecon,jiang2022selfrecon}, lifting single images to high-fidelity clothed humans~\cite{li2025pshuman,saito2019pifu}, or refining surface details via test-time optimization~\cite{kairanda2025thin}.
All remain anchored to parametric kinematics~\cite{ye2025freecloth}, lacking a unified treatment of non-rigid surface motion across time.
Conversely, generic 3D scene flow methods such as RAFT-3D~\cite{teed2021raft}, FlowNet3D~\cite{liu2019flownet3d}, PointPWC-Net~\cite{wu2020pointpwc}, and Self-Mono-SF~\cite{hur2020selfsupervised} natively capture dense temporal dynamics, with recent neural-prior~\cite{li2021nsfp} and ODE-based~\cite{vedder2025eulerflow} formulations further improving generalization.
However, these methods rely on local rigid-body assumptions to regularize the ill-posed matching problem~\cite{lin2025voteflow,li2022rigidflow,lin2024icp}.
Such rigidity priors break down when confronted with highly articulated, non-rigid human movements.
Our \paper bridges these two paradigms, providing a dense human scene flow that unifies skeletal kinematics with fine-grained surface deformation through joint multi-modal estimation.

\paragraph{Self-supervised Motion Learning}
Estimating dense human scene flow without ground truth is a highly ill-posed problem.
To constrain the search space, self-supervised scene flow methods rely on rigidity priors, enforcing them either architecturally~\cite{lin2025voteflow} or through pseudo-label generation~\cite{li2022rigidflow,mittal2020justgowiththeflow}.
Such regularization causes feature tearing and unnatural stiffening on articulated, non-rigid human bodies.
Recent work alternatively scales up training through mixed 3D and weak 2D supervision across diverse synthetic scenes~\cite{liang2025zeroshot,vedder2024zeroflow}, achieving strong zero-shot generalization.
These recipes nonetheless inherit biases toward rigid scenes and degrade on deformable surfaces such as clothing.
A parallel line of work learns humans directly from monocular video, recovering animatable avatars through implicit fields~\cite{weng2022humannerf,jiang2023instantavatar}, scene decomposition~\cite{guo2023vid2avatar}, layered non-rigid rendering~\cite{guo2024reloo}, or 3D Gaussian splatting~\cite{qian2024_3dgsavatar,kocabas2024hugs,guo2025vid2avatarpro}.
These methods target photorealistic appearance rather than dense motion, supervising shape and texture through volume rendering without producing a dense motion field.
\paper takes a different route to self-supervision.
We integrate three classes of physics-inspired priors that govern human motion.
Geometric silhouettes, structural kinematics, and biomechanical dynamics together form a unified paradigm that couples motion estimation across modalities.

\paragraph{Human-centric Motion Benchmarks}
\label{sec:related-benchmark}
Existing benchmarks evaluate different facets of human motion, but none directly measure dense surface dynamics.
Human3.6M~\cite{ionescu2014human3}, MPI-INF-3DHP~\cite{mehta2017monocular}, 3DPW~\cite{vonmarcard2018recovering}, and AMASS~\cite{mahmood2019amass} accurately capture 3D kinematics across lab, in-the-wild, and mocap settings~\cite{shahroudy2016ntu,liu2020ntu,sigal2010humaneva}, leaving the clothed surface unmodeled.
Dense flow benchmarks cover surface motion broadly but not clothed human dynamics specifically.
KITTI~\cite{menze2015object} targets rigid driving scenes, and MPI Sintel~\cite{butler2012naturalistic} evaluates optical flow in animated scenes with generic content~\cite{mayer2016flyingthings3d,wilson2021argoverse2}.
DeformingThings4D~\cite{li20214dcomplete} provides non-rigid ground truth, though its animations are dominated by animals and stylized CG humanoid rigs.
Human-specific datasets incorporate richer body motion while remaining anchored to parametric meshes~\cite{cai2022humman,yu2020humbi}.
SURREAL~\cite{varol2017learning}, AGORA~\cite{patel2021agora}, and M3GYM~\cite{xu2025m3gym} render or fit SMPL bodies with varied clothing and backgrounds, yet evaluate 3D pose and shape rather than surface flow.
BEDLAM~\cite{black2023bedlam} extends this paradigm by rendering physics-simulated garments on moving bodies, but its evaluation still targets SMPL-X body recovery rather than dense surface motion.
4D-DRESS~\cite{wang2024_4ddress} departs from synthetic pipelines~\cite{ma2020cape,bertiche2020cloth3d,zou2023cloth4d}, providing real-world 4D scans of clothing, yet its scale is limited to a small set of subjects and outfits.
\dynact addresses this gap.
Built in a controllable simulator, it provides exact dense flow ground truth for non-rigid human motion across diverse subjects, garments, and actions, enabling rigorous evaluation of surface-level dynamics that existing benchmarks cannot supply.
\section{Methodology}

\subsection{Problem Formulation and Architecture}
\label{sec:formulation}

\paper estimates depth, flow, mask, pose, and camera as joint outputs of a single network.
This reflects the view that they are projections of one underlying physical system, a body in motion through space.
The joint formulation also sidesteps the supervision barrier that direct flow estimation would face.
Real-world per-pixel 3D motion on humans cannot be captured directly.
Marker-based motion capture records only sparse anatomical landmarks, and skin sliding decouples those markers from the underlying body surface.
We therefore cast the problem as joint optimization of \paper under a constraint set distilled from physical priors of human movement.

Let $\mathcal{X} = \{I_i\}_{i \in \mathcal{T}}$ denote a video clip indexed by $\mathcal{T}$, where $I_i$ is the frame at index $i$.
A network $G_\theta$ maps the clip to a sequence of per-frame estimates,
\begin{equation}
    G_\theta : \big\{I_i\big\}_{i \in \mathcal{T}} \;\mapsto\; \big\{\mathcal{Y}_i\big\}_{i \in \mathcal{T}}.
    \label{eq:formulation-forward}
\end{equation}
Each per-frame tuple $\mathcal{Y}_i$ collects the five outputs jointly produced for frame $i$,
\begin{equation}
    \mathcal{Y}_i = \big(\mathbf{D}_i,\; \mathbf{F}_i,\; \mathbf{M}_i,\; \mathbf{P}_i,\; \mathbf{C}_i\big),
    \label{eq:formulation-outputs}
\end{equation}
covering dense depth $\mathbf{D}_i$ at every pixel, forward scene flow $\mathbf{F}_i$ predicting the 3D displacement of every visible point from frame $i$ to frame $i{+}1$, soft human mask $\mathbf{M}_i$ identifying the subject, 3D joint positions $\mathbf{P}_i \in \mathbb{R}^{J \times 3}$ of $J$ skeletal joints, and camera $\mathbf{C}_i$ packaging intrinsics and extrinsics.

\begin{figure}[t]
    \centering
    \includegraphics[width=0.94\linewidth]{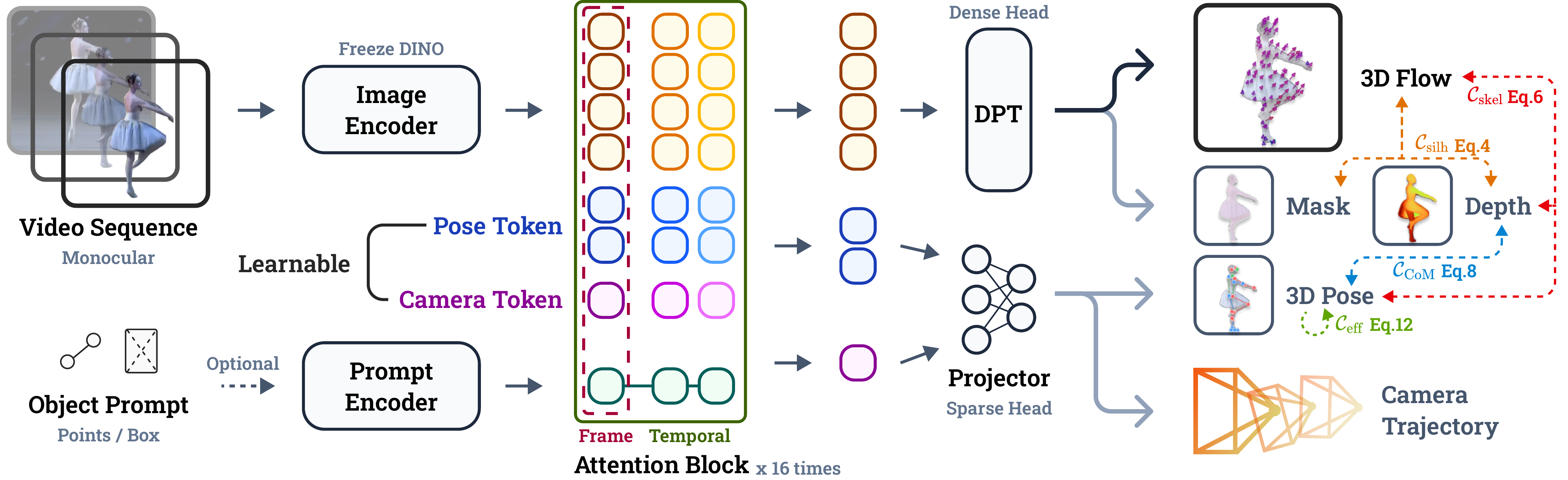}
    \caption{\macaption{\paper architecture.}
    \micaption{We instantiate $G_\theta$ as a single transformer producing the outputs of $\mathcal{Y}$ from a frozen visual substrate.
    Patches and learnable queries (pose, camera) participate in the same attention, so cross-modal coupling exists in the representation before any constraint is computed.}}
    \label{fig:architecture}
    \vspace{-1em}
\end{figure}

We instantiate $G_\theta$ as a single transformer over a frozen DINOv3 backbone~\cite{simeoni2025dinov3} (\cref{fig:architecture}).
Patch features and learnable queries for pose $\mathbf{P}$ and camera $\mathbf{C}$ participate in the same attention, while a DPT-style dense decoder~\cite{ranftl2021vision} produces depth $\mathbf{D}$, flow $\mathbf{F}$, and mask $\mathbf{M}$ from the same patch features.
This shared substrate routes every gradient through one backbone.
Optimization signal from any constraint reaches every output head, instantiating the representation-level coupling within \paper.
Layer counts, query parameterization, and head architecture are deferred to \cref{sec:app_arch}.

Direct regression of flow $\mathbf{F}$ is precluded by the absence of dense supervision on humans.
We instead optimize $\theta$ against differentiable constraints that encode physical priors of human movement:
\begin{equation}
    \theta^\star = \arg\min_\theta \; \mathbb{E}_{\mathcal{X} \sim \mathcal{D}} \left[ \sum_{k} \lambda_k \, \mathcal{C}_k\big(G_\theta(\mathcal{X})\big) \right].
    \label{eq:formulation-objective}
\end{equation}
Each constraint $\mathcal{C}_k \geq 0$ measures how much the network outputs $G_\theta(\mathcal{X})$ violate physical prior $k$, with $\mathcal{C}_k = 0$ corresponding to full satisfaction.
This recasts self-supervised motion estimation as constraint satisfaction over physical priors rather than label regression.
The constraint set partitions into three principled scales.
The geometric constraint acts on per-frame silhouette agreement.
Structural constraints couple skeletal kinematics with surface flow and ground support at each instant.
The biomechanical constraint regularizes multi-frame trajectories against energy-optimal motion templates.
Together, the four constraints engage every output modality, shaping each head from multiple angles and instantiating the supervision-level coupling within \paper.
\cref{sec:loss} specifies the four constraints comprising $\{\mathcal{C}_k\}$, illustrated in \cref{fig:loss}.
\subsection{Self-supervised Constraints from Physical Priors}
\label{sec:loss}

\begin{figure}[t]
    \centering
    \includegraphics[width=0.94\linewidth]{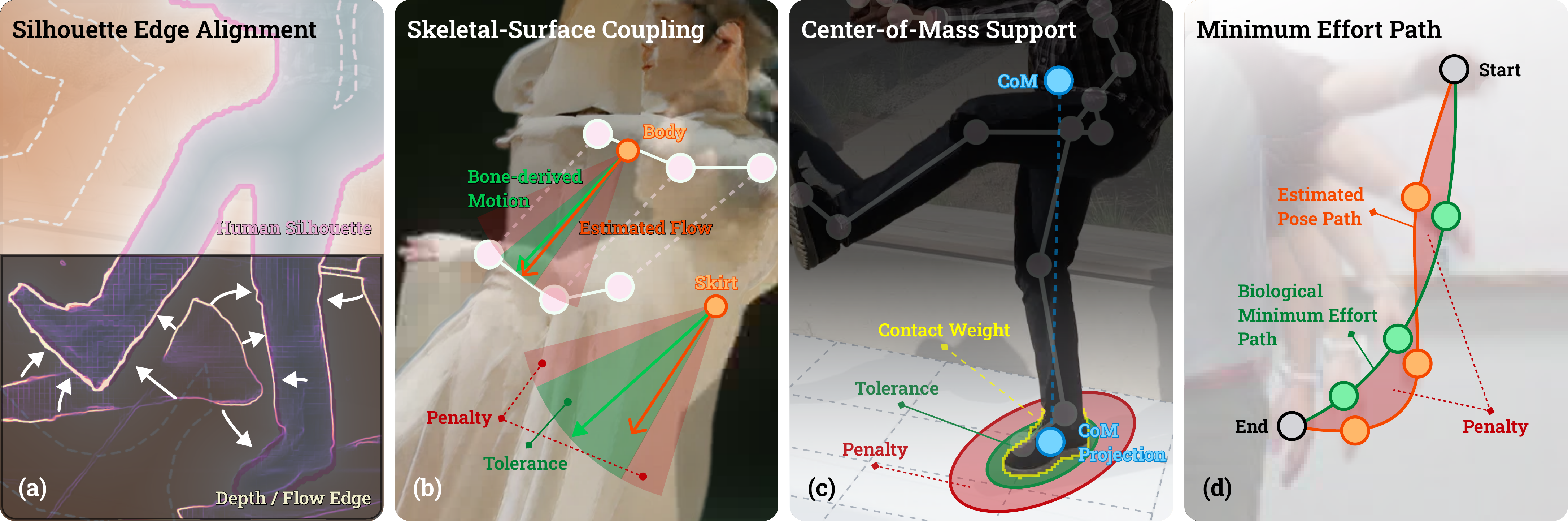}
    \caption{\macaption{Self-supervised constraints from physical priors.}
    \micaption{Four panels illustrate the four constraints that train $G_\theta$ without scene-flow ground truth.
    \emph{(a) Silhouette Edge Alignment} pulls dense edge responses in depth $\mathbf{D}$ and flow $\mathbf{F}$ toward the mask boundary, with a signed distance field penalizing both interior spurious edges and exterior leakage.
    \emph{(b) Skeletal-Surface Coupling} ties surface flow to bone-induced motion via a tolerance margin, allowing loose clothing while anchoring on the skeleton.
    \emph{(c) Center-of-Mass Support} requires the projected mass center to lie within the support polygon formed by ground-contact end-effectors.
    \emph{(d) Minimum Effort Path} regularizes pose trajectories toward the closed-form minimum-jerk arc between endpoint poses.}}
    \label{fig:loss}
    \vspace{-1em}
\end{figure}

\cref{fig:loss} illustrates the four constraints described below.
Throughout, the mask $\mathbf{M}$ acts as a subject selector rather than a measurement target.
It provides the silhouette boundary $\partial\mathbf{M}$, the foreground indicator $\mathbb{1}_\mathbf{M}$, or the admissible region within which other constraints apply.

\paper couples modalities at two complementary levels.
The architecture in \cref{sec:formulation} couples them at the representation level.
As illustrated by the dashed lines in \cref{fig:architecture}, the four constraints below couple them at the supervisory level, so every output head is shaped by gradients from at least two of them.

\paragraph{Geometric: Silhouette Edge Alignment}
Discontinuities in dense estimates should align with the visible body boundary.
A reliable depth field $\mathbf{D}$ has its sharp gradients at body edges, not in the interior.
A reliable flow field $\mathbf{F}$ changes sharply across the moving subject's outline, not within a uniform shirt.
We measure edge response as the gradient norm of each modality, $\|\nabla \mathbf{D}\|_2$ and $\|\nabla \mathbf{F}\|_2$.
We then weight the edge response by distance to the mask boundary $\partial\mathbf{M}$.
The signed distance field $\text{SDF}_{\partial\mathbf{M}}$ is positive outside the mask and negative inside.
Its absolute value is zero on the boundary and grows in either direction.
We index pixels as $\mathbf{u} \in \Omega$ throughout, where $\Omega$ is the image grid.
All dense fields (depth $\mathbf{D}$, flow $\mathbf{F}$, mask $\mathbf{M}$) take values over $\Omega$:
\begin{align}
    \mathcal{C}_\text{silh}(\mathbf{D}, \mathbf{F}) &= \frac{1}{|\Omega|} \sum_{\mathbf{u} \in \Omega} \big( \|\nabla \mathbf{D}(\mathbf{u})\|_2 + \|\nabla \mathbf{F}(\mathbf{u})\|_2 \big) \cdot \big| \text{SDF}_{\partial\mathbf{M}}(\mathbf{u}) \big|.
    \label{eq:loss-silh}
\end{align}
Symmetric penalty on both sides of $\partial\mathbf{M}$ pulls interior spurious edges out toward the boundary and exterior leakage in (\cref{fig:loss}{\small a}).
The constraint thus couples depth $\mathbf{D}$ and flow $\mathbf{F}$ through a shared geometric reference provided by mask $\mathbf{M}$.

\paragraph{Structural: Skeletal-Surface Coupling}
Surface motion is not free.
Every surface point sits within a few centimeters of an underlying bone, and its first-order motion is dominated by that bone's kinematics, even when clothing deforms non-rigidly.
We encode this by lifting each foreground pixel into a 3D surface point through depth unprojection.
The point is then assigned to its nearest bone, and the predicted flow $\mathbf{F}$ at that pixel is compared against the bone-induced motion derived from the joint trajectory.
Let $b^* = (j_1^*, j_2^*)$ index the nearest bone, with $\alpha^* \in [0,1]$ marking its barycentric position along that bone.
The bone-induced motion is a linear interpolation of the two endpoint joint motions.
The actual flow is penalized only when its deviation exceeds an anatomical tolerance $\rho(\mathbf{u})$ that grows with distance from the skeleton:
\begin{align}
    \mathbf{m}_\text{bone}(\mathbf{u}) &= (1-\alpha^*)\,\Delta\mathbf{P}_{j_1^*} + \alpha^*\,\Delta\mathbf{P}_{j_2^*}, \label{eq:loss-skel-bone}\\
    \mathcal{C}_\text{skel}(\mathbf{F}, \mathbf{D}, \mathbf{P}) &= \frac{1}{N_\mathbf{M}} \sum_{\mathbf{u}} \mathbb{1}_\mathbf{M}(\mathbf{u}) \cdot \big[\, \|\mathbf{F}(\mathbf{u}) - \mathbf{m}_\text{bone}(\mathbf{u})\|_2 - \rho(\mathbf{u}) \,\big]_+, \label{eq:loss-skel}
\end{align}
where $\Delta\mathbf{P}_j$ is the per-joint motion vector and $N_\mathbf{M} = \sum_\mathbf{u} \mathbb{1}_\mathbf{M}(\mathbf{u})$ normalizes by foreground area (\cref{fig:loss}{\small b}).
The tolerance follows a piecewise-linear form $\rho(\mathbf{u}) = \rho_\text{min} + \alpha \cdot d_\text{bone}(\mathbf{u})$.
Here $d_\text{bone}(\mathbf{u})$ is the 3D distance from the unprojected surface point to its nearest bone.
The form admits clothing dynamics.
Distant points get a wide margin, while on-skin points stay close to the skeleton.
Because the unprojection step involves depth $\mathbf{D}$, the constraint backpropagates through depth, flow, and pose simultaneously, coupling all three.

\paragraph{Structural: Center-of-Mass Support}
A standing or stepping body must keep its mass center over a base of support.
The predicted center of mass $\boldsymbol{\mu}$ is a weighted average of joint positions, using anatomical mass distribution constants $w_j$ that vary by body segment.
Ground contact points are read off the pose $\mathbf{P}$ with depth-based proximity.
Their convex hull on the ground plane defines the dynamic support polygon $\mathcal{S}(\mathbf{P}, \mathbf{D})$.
The constraint penalizes the signed distance by which the projected mass center $\boldsymbol{\mu}_{xy}$ leaves this polygon.
Inside the polygon counts as zero violation by construction:
\begin{align}
    \boldsymbol{\mu} &= \sum_{j=1}^{J} w_j\, \mathbf{P}_j, \label{eq:loss-com-mu}\\
    \mathcal{C}_\text{CoM}(\mathbf{P}, \mathbf{D}) &= \big[\, \text{SDF}\big(\boldsymbol{\mu}_{xy},\, \partial \mathcal{S}(\mathbf{P}, \mathbf{D})\big) \,\big]_+. \label{eq:loss-com}
\end{align}
The polygon construction depends on depth $\mathbf{D}$ for ground geometry and on mask $\mathbf{M}$ to restrict candidate contacts to foreground regions.
Full details are in \cref{sec:app_losses}.
Beyond stability, the constraint signals when a depth estimate places the body floating above or sinking below the ground (\cref{fig:loss}{\small c}).
It thus couples pose and depth through a single physical principle.

\paragraph{Biomechanical: Minimum Effort Path}
Voluntary human movement between two postures does not take an arbitrary detour.
Decades of motor control research show that point-to-point movement follows energy-minimizing trajectories, formalized as the minimum-jerk principle~\cite{flash1985coordination, hogan1984}.
Closely related formulations include minimum torque-change~\cite{uno1989} and optimal feedback control~\cite{todorov2002}.
We encode this canonical model directly.
For a temporal window $\mathcal{W}$ with endpoint frames $i_0$ and $i_T$, we map the endpoint poses $\mathbf{P}_{i_0}$ and $\mathbf{P}_{i_T}$ into joint-angle space through inverse kinematics, yielding $\mathbf{q}_{i_0}$ and $\mathbf{q}_{i_T}$.
We then interpolate $\mathbf{q}^\star_i$ along the closed-form minimum-jerk arc between them, and forward-kinematic the result back to position space as the reference pose $\mathbf{P}^\star_i$:
\begin{align}
    \mathbf{q}_{i_0} &= \text{IK}\big(\text{sg}(\mathbf{P}_{i_0})\big), \quad \mathbf{q}_{i_T} = \text{IK}\big(\text{sg}(\mathbf{P}_{i_T})\big), \label{eq:loss-eff-ik}\\
    \mathbf{q}^\star_i &= \mathbf{q}_{i_0} + (\mathbf{q}_{i_T} - \mathbf{q}_{i_0}) \cdot \phi(\tau_i), \quad \phi(\tau) = 10\tau^3 - 15\tau^4 + 6\tau^5, \label{eq:loss-eff-jerk}\\
    \mathbf{P}^\star_i &= \text{FK}(\mathbf{q}^\star_i), \label{eq:loss-eff-fk}\\
    \mathcal{C}_\text{eff}\big(\{\mathbf{P}_i\}_{i \in \mathcal{W}}\big) &= \frac{1}{|\mathcal{W}|} \sum_{i \in \mathcal{W}} \big[\, \|\mathbf{P}_i - \mathbf{P}^\star_i\|_2 - \rho_\text{eff} \,\big]_+, \label{eq:loss-eff}
\end{align}
where $\tau_i = (i - i_0)/(i_T - i_0)$ and $\text{sg}(\cdot)$ is stop-gradient.
The 5th-order polynomial $\phi$ is the closed-form minimizer of integrated jerk between fixed endpoints.
The constraint penalizes the deviation between the estimated pose $\mathbf{P}_i$ and the minimum-jerk reference $\mathbf{P}^\star_i$, beyond a tolerance margin $\rho_\text{eff}$.
Stop-gradient on the endpoints prevents the network from co-adapting boundary poses to the reference, anchoring regularization on intermediate dynamics (\cref{fig:loss}{\small d}).
Although the constraint nominally reads only pose $\mathbf{P}$, gradient flows back through the shared trunk and reaches every modality.

\paragraph{Anchoring and consistency}
The four constraints above do not exhaust the full optimization.
To prevent degenerate minima admitted by self-supervision alone, we apply a margin-based distillation anchor on depth $\mathbf{D}$ and pose $\mathbf{P}$ against off-the-shelf teachers~\cite{bochkovskii2024depthpro,yang2026sam3dbody}, plus a single-camera consistency term on camera $\mathbf{C}$.
These auxiliary terms support the optimization but do not define the physical-prior framework.
Their formulation is deferred to \cref{sec:app_anchor}.
\cref{sec:experiments} evaluates the resulting estimator across multiple datasets.
\section{\dynact Benchmark}
\label{sec:dynact}

\dynact bypasses the composite path used by prior synthetic benchmarks (\cref{sec:related-benchmark}) by tracking the physical surface through the renderer's skinning stage.
We record the world-space position of every skinned mesh vertex at floating-point precision.
A custom G-buffer pass then propagates this vertex motion to every pixel through triangle identifiers and barycentric coordinates captured at rasterization.
With 24K vertices per character, the mesh resolution exceeds the pixel grid at all camera distances we render, so every pixel maps to at least three host vertices.
The resulting per-pixel flow is exact rather than interpolated from sparse samples.

\dynact is built on 10 characters from MetaHuman~\cite{epic2024metahuman} framework, generated within Unreal Engine~\cite{epic2022unreal}, spanning genders, ethnicities, body types, and heights.
Each character is paired with 10 garment configurations, from tight-fitted to loose-flowing.
Every subject performs 10 action categories of $\sim$15\,s each, sourced from the Mixamo animation library~\cite{mixamo} and retargeted to the MetaHuman rig.
Actions cover near-static postures, locomotion, and highly articulated movements such as dancing and martial arts.
We sample 100 motion sequences from the $10^3$ candidate combinations.
Each sampled motion is rendered at all 8 synchronized viewpoints across 3 environments spanning indoor and outdoor scenes.
The result is 800 test sequences totaling $\sim$720K frames at $1920 \times 1080$ and 60\,fps.
\cref{tab:benchmark-comparison} shows representative frames.
We position \dynact entirely as a test set.
Every method reports results under zero-shot transfer, trained on external data and evaluated on \dynact without exposure to its subjects, actions, scenes, or viewpoints.

\begin{table}[t]
    \caption{\macaption{\dynact vs existing human motion and scene flow benchmarks, and representative frames.}
    \micaption{\dynact pairs simulated clothed humans with vertex-level flow ground truth, while prior synthetic benchmarks either composite 2D flow with depth or omit flow entirely. ``---'' indicates unavailable. Right: representative frames showing diverse subjects, garments, and actions under different viewpoints.}}
    \label{tab:benchmark-comparison}
    \centering
    \begin{minipage}[c]{0.68\textwidth}
        \centering
        \centering
\resizebox{\linewidth}{!}{%
    \begin{tabular}{lccccr}
        \toprule
        \textbf{Dataset}                                  & \textbf{Domain} & \textbf{Source} & \textbf{Clothing}             & \textbf{4D Flow GT}            & \textbf{Frames}   \\
        \midrule
        Human3.6M~\cite{ionescu2014human3}                & Human           & Real            & ---                           & ---                            & 3.6M                \\
        3DPW~\cite{vonmarcard2018recovering}              & Human           & Real            & SMPL fits                     & ---                            & 51K                 \\
        AMASS~\cite{mahmood2019amass}                     & Human           & Mocap           & SMPL body                     & ---                            & ---                 \\
        \midrule
        KITTI~\cite{menze2015object}                      & Scene           & Real            & ---                           & LiDAR                          & 400                 \\
        MPI Sintel~\cite{butler2012naturalistic}          & Scene           & Syn.            & ---                           & Composite                      & 1K                  \\
        DeformingThings4D~\cite{li20214dcomplete}         & Animal          & Syn.            & ---                           & Vertex                         & 500K                \\
        \midrule
        AGORA~\cite{patel2021agora}                       & Human           & Syn.            & Rigged                        & ---                            & 17K                 \\
        M3GYM~\cite{xu2025m3gym}                          & Human           & Syn.            & SMPL fits                     & ---                            & 4M                  \\
        BEDLAM~\cite{black2023bedlam}                     & Human           & Syn.            & Simulated                     & Composite                      & 380K                \\
        \textbf{\dynact} {\scriptsize (\textbf{Ours})}      & \textbf{Human}  & \textbf{Syn.}   & \textbf{Simulated}            & \textbf{Vertex}                & \textbf{720K}       \\
        \bottomrule
    \end{tabular}
}
    \end{minipage}
    \hfill
    \begin{minipage}[c]{0.31\textwidth}
        \centering
        \includegraphics[width=0.94\linewidth]{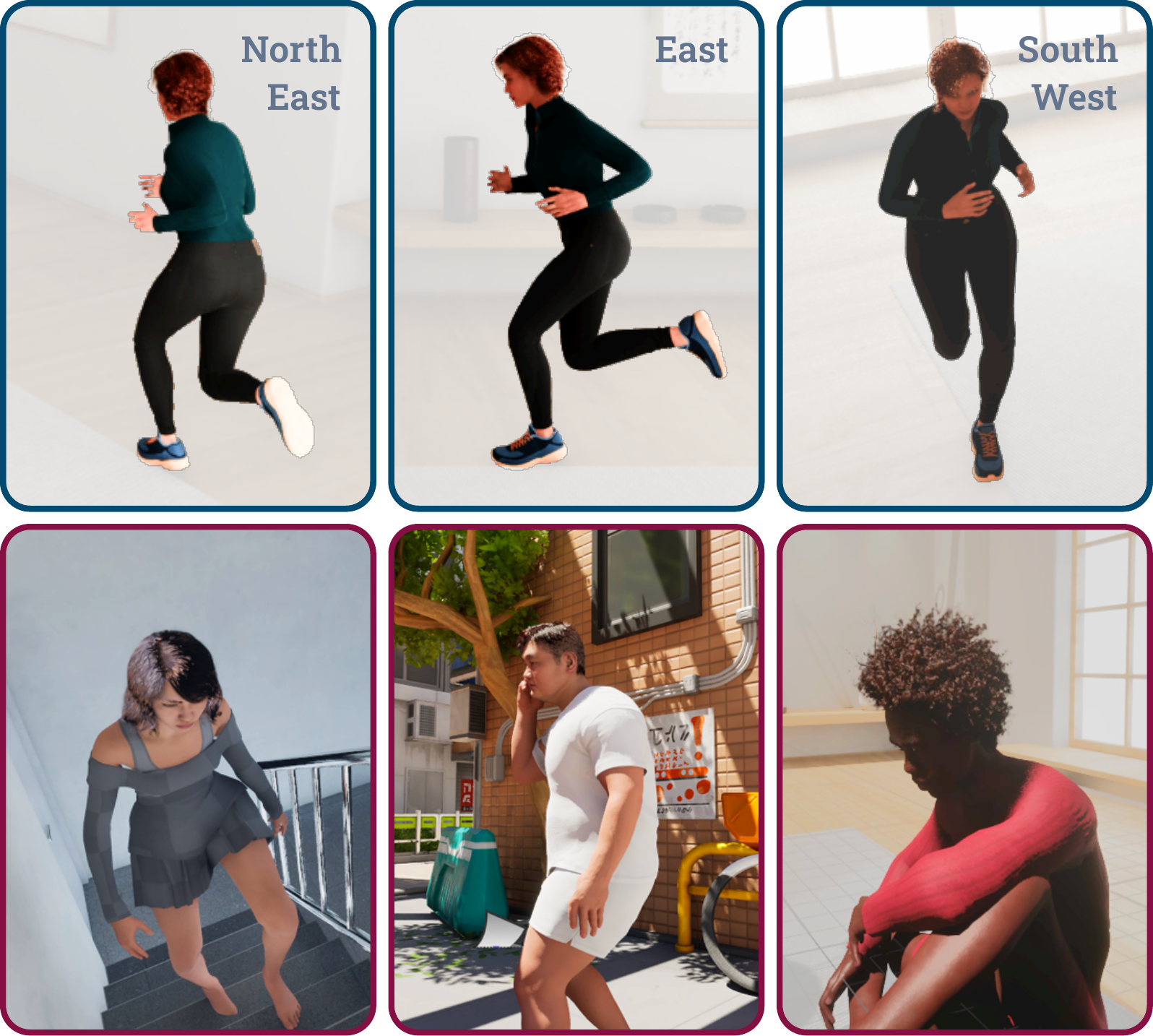}
    \end{minipage}
    \vspace{-1.4em}
\end{table}
\section{Experiments}
\label{sec:experiments}

We pretrain \paper on the full Human3.6M~\cite{ionescu2014human3} dataset to learn cross-modal human priors.
Since we do not evaluate on Human3.6M, all seven subjects (S1, S5, S6, S7, S8, S9, S11) contribute to pretraining.
To establish in-domain evaluation on Fit3D~\cite{fieraru2021aifit} and 3DPW~\cite{vonmarcard2018recovering}, we then fine-tune on their respective train splits.
For Fit3D we use S3, S4, S5, S7, S8, S9 for training and S10, S11 for testing; for 3DPW we use the official partition.
\dynact serves as an out-of-domain benchmark, evaluated under zero-shot inference.
Enabled by the proposed self-supervised paradigm, \paper trains on RGB video alone, without any modality ground truth from these datasets.
We compare against representative methods spanning multiple categories of monocular motion estimation, with task-specific baselines and citations in the corresponding tables.
We implement \paper in PyTorch with mixed-precision training over $4 \times 10^5$ iterations on 6 NVIDIA H100 GPUs.

\subsection{Scene Flow Estimation}
\label{sec:exp-sf}

We adopt Fit3D and \dynact as the benchmarks for this section.
Fit3D combines professional motion capture with multi-view fusion to produce high-fidelity SMPL-X annotations, and all subjects wear tight-fitting attire so that the same vertex remains stable across consecutive frames; per-vertex displacement therefore serves as faithful scene flow ground truth.
\dynact provides per-pixel scene flow ground truth and includes garment-fold dynamics, further probing how \paper handles fine-grained surface deformation on clothed humans.
3DPW is excluded as a scene flow source because its SMPL annotations are not directly mocap-captured and its subjects are not in tight-fitting attire; we therefore use it only for pose and depth evaluation.

We compare \paper against six representative baselines, two from each of three categories that produce dense scene flow.
\emph{(i) Scene flow}~\cite{liang2025zeroshot,hur2020selfsupervised} regresses 3D motion vectors from RGB pairs.
\emph{(ii) Optical flow lifting}~\cite{bochkovskii2024depthpro,huang2025hmore,teed2020raft} lifts off-the-shelf 2D optical flow into 3D using estimated depth.
\emph{(iii) Mesh residual}~\cite{yang2026sam3dbody,shin2024wham} reads vertex displacement from a fitted body model and propagates it to dense pixels via nearest-neighbor.

\paragraph{Quantitative Results}
As shown in \cref{tab:sf-main}, \paper achieves the lowest 3D end-point error on both datasets across every metric.
On Fit3D, the closest competitor is the Depth Pro + H-MoRe pipeline at $38.3$\,mm, which \paper improves to $24.2$\,mm, a $37\%$ reduction.
The same baseline on \dynact reaches $42.2$\,mm, against \paper's $28.2$\,mm, a $33\%$ reduction under out-of-domain transfer.
The gap widens for generic scene flow methods that lack human-specific priors.
ZeroMSF~\cite{liang2025zeroshot} and Self-Mono-SF~\cite{hur2020selfsupervised} both leave EPE above $50$\,mm even after Fit3D fine-tuning, since their rigid-motion regularizers do not transfer to articulated bodies.

\begin{table}[t]
    \caption{\macaption{Scene flow estimation on Fit3D and \dynact.}
    \micaption{End-Point-Error (EPE) in millimeters; 1$-$Cos is one minus cosine similarity between predicted and ground-truth flow vectors; Acc.S and Acc.R are the fractions of pixels within $50$\,mm and $100$\,mm of the ground-truth flow.
    Best results in \best, second-best \second.}}
    \label{tab:sf-main}
    \centering
\resizebox{0.98\textwidth}{!}{
    \small
    \begin{tabular}{llcccccccc}
        \toprule
        \multicolumn{2}{c}{\textbf{Methods}}
            & \multicolumn{4}{c}{\textbf{Fit3D}~\cite{fieraru2021aifit} {\scriptsize \textbf{In-Domain}}}
            & \multicolumn{4}{c}{\textbf{\dynact} {\scriptsize \textbf{Out-of-Domain}}}
        \\
        \cmidrule(lr){1-2} \cmidrule(lr){3-6} \cmidrule(lr){7-10}
        \textbf{Name} & \textbf{Supervision}
        & \textbf{EPE}\(\downarrow\) {\scriptsize{mm}} & \textbf{1$-$Cos}\(\downarrow\) & \textbf{Acc.S}\(\uparrow\)& \textbf{Acc.R}\(\uparrow\)
        & \textbf{EPE}\(\downarrow\) {\scriptsize{mm}} & \textbf{1$-$Cos}\(\downarrow\) & \textbf{Acc.S}\(\uparrow\)& \textbf{Acc.R}\(\uparrow\)
        \\
        \midrule
        \textbf{\paper{}} {\scriptsize \textbf{Ours}}                    & {\scriptsize Self Sup.}    & \textbf{24.2}     & \textbf{0.11}     & \textbf{0.88}     & \textbf{0.92}    & \textbf{28.2}     & \textbf{0.14}     & \textbf{0.82}     & \textbf{0.88}     \\
        ZeroMSF$^\ast$~\cite{liang2025zeroshot}                            & {\scriptsize Geom. Consist.}      & 59.9              & 0.33              & 0.41              & 0.48             & 78.5              & 0.42              & 0.33              & 0.42              \\
        Self-Mono-SF$^\ast$~\cite{hur2020selfsupervised}                   & {\scriptsize Geom. Consist.}      & 50.2              & 0.31              & 0.38              & 0.40             & 68.0              & 0.39              & 0.28              & 0.32              \\
        \midrule
        Depth Pro + H-MoRe~\cite{bochkovskii2024depthpro, huang2025hmore}  & {\scriptsize Self Sup.}    & \underline{38.3}  & \underline{0.18}  & \underline{0.81}  & \underline{0.88} & \underline{42.2}  & \underline{0.20}  & \underline{0.72}  & \underline{0.81}  \\
        Depth Pro + RAFT~\cite{bochkovskii2024depthpro, teed2020raft}      & {\scriptsize Geom. Consist.}      & 85.3              & 0.41              & 0.35              & 0.47             & 122.1             & 0.58              & 0.16              & 0.27              \\
        \midrule
        SAM 3D Body~\cite{yang2026sam3dbody}                        & {\scriptsize N/A}       & 49.6              & 0.26              & 0.78              & 0.82             & 56.8              & 0.25              & 0.66              & 0.70              \\
        WHAM~\cite{shin2024wham}                                    & {\scriptsize Sup.}       & 57.2              & 0.33              & 0.58              & 0.74             & 62.5              & 0.32              & 0.58              & 0.62              \\
        \bottomrule
    \end{tabular}
}
    \vspace{-1em}
\end{table}

\begin{figure}[t]
    \centering
    \includegraphics[width=0.94\linewidth]{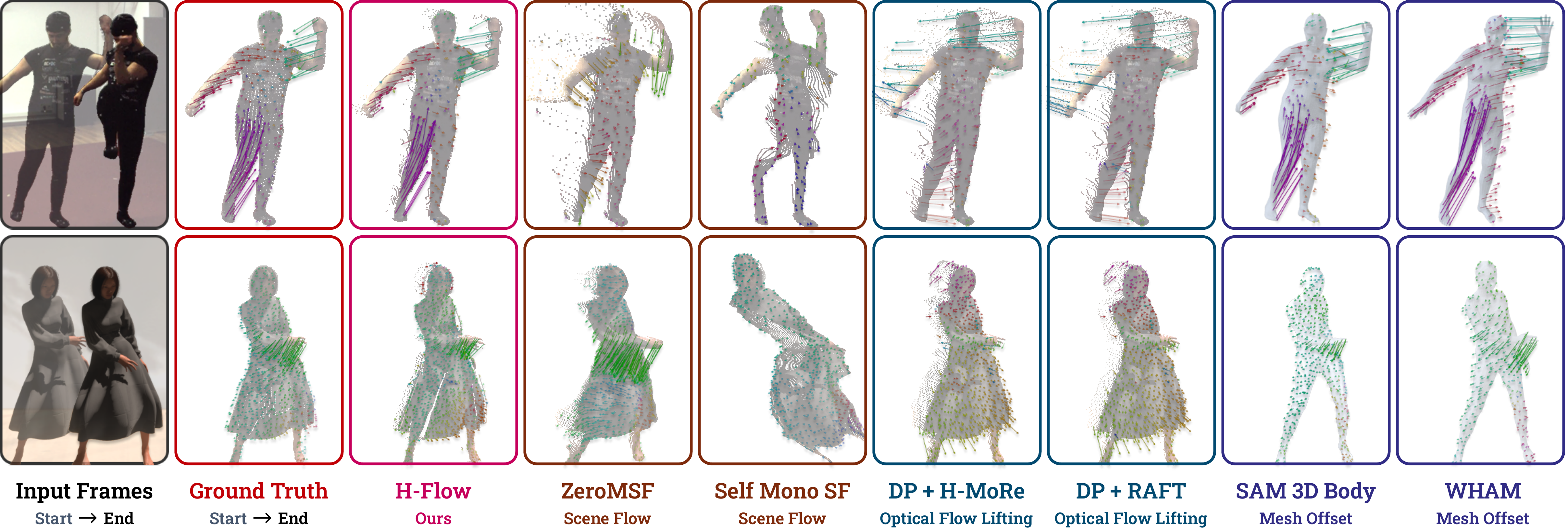}
    \caption{\macaption{Qualitative scene flow comparison.}
    \micaption{\emph{(top)} Fit3D \emph{\footnotesize [in-domain]}, \emph{(bottom)} \dynact \emph{\footnotesize [out-of-domain]}.
    Scene flow baselines smear on articulated limbs and clothing. Lifting pipelines accumulate edge artifacts at silhouette boundaries. Mesh residual methods recover skeletal motion but flatten surface dynamics on garments.
    \emph{A higher-resolution version is provided in the supplementary material.}
    }}
    \label{fig:sf-qualitative}
    \vspace{-1.8em}
\end{figure}

\paragraph{Qualitative Results}
\cref{fig:sf-qualitative} visualizes flow predictions on representative sequences from both datasets.
Generic scene flow baselines produce coherent motion on torsos but smear at limb tips and clothing edges, where rigid-motion priors break.
Pipelines that composite depth with optical flow accumulate edge artifacts at the silhouette boundary, exactly the regions where physical priors anchor \paper's predictions.
Mesh residual methods preserve global pose but flatten secondary surface dynamics, missing the swirl and compression that distinguish clothed-body motion.
\paper recovers fine-grained surface flow that aligns with both skeletal kinematics and clothing deformation.

\subsection{Companion Modality Estimation}
\label{sec:exp-companion}

\paragraph{Pose Estimation}
To verify that joint estimation does not compromise companion modality quality, we compare \paper's pose head against image-based pose estimators and 2D-to-3D lifting methods on Fit3D and 3DPW in \cref{tab:companion}.
\paper achieves the lowest MPJPE on both datasets, improving over dedicated estimators including SAM 3D Body~\cite{yang2026sam3dbody}, CameraHMR~\cite{patel2025camerahmr}, and PromptHMR~\cite{wang2025prompthmr}.
This advantage concentrates in the absolute joint position error.
Joint estimation provides \paper's pose head with the metric depth required for absolute world-space localization, a signal that dedicated pose estimators must infer indirectly from monocular cues.
On PA-MPJPE, where global scale and rotation are removed by Procrustes alignment, methods such as HiPART~\cite{zheng2025hipart} and MotionAGFormer~\cite{mehraban2024motionagformer} remain competitive.
These take 2D keypoints as input and benefit from the much larger pose datasets available for training relative-joint configuration.
\paper's pose head sits competitively in this regime while serving as one of five jointly trained outputs, rather than a specialized pose pipeline.

\paragraph{Depth Estimation}
Ground-truth depth on Fit3D and 3DPW is derived by rasterizing the official SMPL-X fits.
The resulting per-pixel depth covers the body surface only, while clothing and scene depth are not annotated.
We evaluate every method within this body region uniformly, shown in \cref{tab:companion}.
This region happens to align with \paper's design, where the depth head is consumed only by priors that operate on the body surface.
Within this region, \paper substantially improves over generic monocular metric depth predictors that estimate depth uniformly over the full scene.
On Fit3D, MAE drops from $234.3$\,mm for Depth Pro~\cite{bochkovskii2024depthpro} to $124.5$\,mm for \paper, a $47\%$ reduction.
The same pattern holds across all metrics on both Fit3D and 3DPW.
Generic predictors allocate model capacity across backgrounds and clothing alike.
\paper concentrates capacity on the human surface that matters for downstream scene flow.

\begin{table}[t]
  \centering
    \caption{\macaption{Companion modality estimation on 3DPW and Fit3D.}
    \micaption{\emph{(a)} Pose estimation, mean per-joint position error (MPJPE) and Procrustes-aligned MPJPE (PA) in mm. Methods marked with $^\dagger$ take 2D keypoints as input.
    \emph{(b)} Depth estimation within the SMPL-X body region, mean absolute error (MAE) in mm and SiLog the scale-invariant log error.
    Best results in \best, second-best \second.}}
  \label{tab:companion}
  \footnotesize
  \setlength{\tabcolsep}{4pt}
  \renewcommand{\arraystretch}{1.0}

  \begin{minipage}[t]{0.48\linewidth}
    \centering
    \resizebox{\linewidth}{!}{\begin{tabular}{lcccc}
  \toprule
  \multirow{2}{*}{\textbf{Methods}}
      & \multicolumn{2}{c}{\textbf{3DPW}}
      & \multicolumn{2}{c}{\textbf{Fit3D}} \\
  \cmidrule(lr){2-3} \cmidrule(lr){4-5}
      & \textbf{MPJPE}\(\downarrow\) {\scriptsize mm} & \textbf{PA}\(\downarrow\) {\scriptsize mm}
      & \textbf{MPJPE}\(\downarrow\) {\scriptsize mm} & \textbf{PA}\(\downarrow\) {\scriptsize mm} \\
  \midrule
  \textbf{\paper (Ours)}                                  & \textbf{91.5}     & 48.7              & \textbf{112.8}    & 55.4              \\
  \midrule
  HiPART$^\dagger$~\cite{zheng2025hipart}                   & 112.4             & \textbf{44.5}     & 138.5             & \textbf{51.2}     \\
  MotionAGFormer$^\dagger$~\cite{mehraban2024motionagformer} & 108.6            & \underline{45.8}  & 135.8             & \underline{52.6}  \\
  MotionBERT$^\dagger$~\cite{zhu2023motionbert}             & 110.2             & 46.3              & 137.4             & 53.2              \\
  \midrule
  SAM 3D Body~\cite{yang2026sam3dbody}                      & \underline{95.0}  & 45.0              & \underline{116.0} & 54.0              \\
  CameraHMR~\cite{patel2025camerahmr}                       & 98.3              & 46.5              & 120.4             & 55.8              \\
  PromptHMR~\cite{wang2025prompthmr}                        & 102.7             & 48.9              & 124.1             & 57.2              \\
  \bottomrule
\end{tabular}}
  \end{minipage}\hfill
  \begin{minipage}[t]{0.51\linewidth}
    \centering
    \resizebox{\linewidth}{!}{\begin{tabular}{lcccc}
  \toprule
  \multirow{2}{*}{\textbf{Methods}}
      & \multicolumn{2}{c}{\textbf{3DPW}}
      & \multicolumn{2}{c}{\textbf{Fit3D}} \\
  \cmidrule(lr){2-3} \cmidrule(lr){4-5}
      & \textbf{MAE}\(\downarrow\) {\scriptsize mm} & \textbf{SiLog}\(\downarrow\)
      & \textbf{MAE}\(\downarrow\) {\scriptsize mm} & \textbf{SiLog}\(\downarrow\) \\
  \midrule
  \textbf{\paper{} (Ours)}                          & \textbf{158.6}    & \textbf{7.2}      & \textbf{124.5}    & \textbf{2.9}      \\
  \midrule
  Depth Pro~\cite{bochkovskii2024depthpro}          & 527.7             & 12.4              & 234.3             & 4.4               \\
  Depth Anything V2~\cite{yang2024depthv2}  & \underline{485.3} & \underline{11.8}  & \underline{218.7} & \underline{4.1}   \\
  MoGe-2~\cite{wang2025moge2}                       & 452.1             & 11.3              & 205.4             & 3.9               \\
  Marigold~\cite{ke2024marigold}                    & 498.2             & 12.0              & 225.6             & 4.2               \\
  ZoeDepth~\cite{bhat2023zoedepth}                  & 512.8             & 12.2              & 228.9             & 4.3               \\
  \bottomrule
\end{tabular}}
  \end{minipage}
  \vspace{-1.2em}
\end{table}

\subsection{Ablations}
\label{sec:exp-ablations}

We ablate each physical prior and the distillation anchor in turn (\cref{tab:ablation}), retraining \paper with the same setup minus one component.

\begin{table}[t]
    \caption{\macaption{Ablation of physical priors and the distillation anchor.}
    \micaption{Each row removes one component. Scene flow uses EPE (mm) and 1$-$Cos on Fit3D and \dynact; pose uses MPJPE and PA (mm) on 3DPW; depth uses MAE (mm) and scale-invariant SiLog on Fit3D. Lower is better for all metrics.}}
    \label{tab:ablation}
    \centering
\resizebox{0.98\textwidth}{!}{
    \small
\begin{tabular}{lcccccccc}
    \toprule
    \multirow{2}{*}{\textbf{Methods}}
    & \multicolumn{2}{c}{\textbf{Flow {\scriptsize on Fit3D}}}
    & \multicolumn{2}{c}{\textbf{Flow {\scriptsize on \dynact}}}
    & \multicolumn{2}{c}{\textbf{Pose {\scriptsize on 3DPW}}}
    & \multicolumn{2}{c}{\textbf{Depth {\scriptsize on Fit3D}}}
    \\
    \cmidrule(lr){2-3} \cmidrule(lr){4-5} \cmidrule(lr){6-7} \cmidrule(lr){8-9}
    & \textbf{EPE}\(\downarrow\) {\scriptsize mm} & \textbf{1$-$Cos}\(\downarrow\)
    & \textbf{EPE}\(\downarrow\) {\scriptsize mm} & \textbf{1$-$Cos}\(\downarrow\)
    & \textbf{MPJPE}\(\downarrow\) {\scriptsize mm} & \textbf{PA}\(\downarrow\) {\scriptsize mm}
    & \textbf{MAE}\(\downarrow\) {\scriptsize mm} & \textbf{SiLog}\(\downarrow\)
    \\
    \midrule
    \paper                                & \textbf{24.2} & \textbf{0.11} & \textbf{28.2} & \textbf{0.14} & \textbf{91.5}  & \textbf{48.7} & \textbf{124.5} & \textbf{2.9} \\
    \midrule
    w/o $\mathcal{C}_\text{silh}$                & 32.9          & 0.15          & 41.8          & 0.22          & 92.6           & 49.3          & 165.0          & 3.3          \\
    w/o $\mathcal{C}_\text{skel}$                & 48.4          & 0.22          & 63.6          & 0.21          & 98.1           & 52.3          & 164.8          & 3.1          \\
    w/o $\mathcal{C}_\text{CoM}$                 & 26.2          & 0.12          & 30.9          & 0.16          & 104.6          & 54.5          & 133.0          & 3.3          \\
    w/o $\mathcal{C}_\text{eff}$                 & 42.7          & 0.14          & 44.5          & 0.18          & 102.8          & 52.8          & 126.1          & 3.0          \\
    \midrule
    w/o distillation anchor                       & 55.1          & 0.13          & 44.8          & 0.17          & 183.3          & 47.2          & 276.8          & 2.8          \\
    \bottomrule
\end{tabular}}
    \vspace{-1em}
\end{table}

\paragraph{Cross-modal effects}
Each prior degrades a different lead modality.
$\mathcal{C}_\text{silh}$ mainly sharpens depth, with indirect gains on scene flow at edge regions.
$\mathcal{C}_\text{skel}$ is the dominant scene flow regularizer, removing it doubles EPE on Fit3D and more than doubles it under out-of-domain transfer to \dynact.
$\mathcal{C}_\text{CoM}$ leaves the largest mark on pose, where the support-polygon constraint contributes physical plausibility.
$\mathcal{C}_\text{eff}$ acts on pose trajectories yet propagates through skeletal coupling.
Pose error grows modestly while Fit3D scene flow EPE rises by $76\%$, consistent with \paper's design where every prior reaches multiple heads through the shared backbone.

\paragraph{Distillation as anthropometric anchor}
Removing the distillation anchor splits the metrics into two regimes.
Absolute errors (EPE, MPJPE, MAE) double or triple, while scale-invariant metrics (PA-MPJPE, SiLog) remain at the full-model level.
The four physical priors are scale-invariant by construction; none constrains the global metric scale of the world.
The distillation anchor injects this missing degree of freedom during training through noisy teacher predictions, and the network internalizes it as a learned anthropometric prior.
At inference time the prior is self-contained, and \paper's out-of-domain \dynact results (\cref{tab:sf-main}) confirm that this anthropometric scale transfers across distributions without runtime calibration.
\section{Conclusion}
\label{sec:conclusion}

In this paper, we show that physical priors on the body's surface and skeleton supervise dense scene flow from monocular video without any 4D ground truth.
A learned anthropometric prior, internalized at training, fixes the global metric scale at inference.
\paper sets a new state of the art on absolute scene flow across three datasets while preserving competitive companion modalities at inference without runtime calibration.
A real limitation remains.
The four physical priors are scale-invariant by construction, and \paper relies on a distillation anchor against noisy teacher predictions to fix global metric scale during training.
A fully self-supervised path to anthropometric grounding stays open.
Beyond the human body, dense flow over the contact between bodies and objects is the natural next step.
Extending \paper to human-object interaction is a direct application of its existing architecture, with new physical priors needed for contact, grasp, and release dynamics.

\newpage
{
\small
\bibliographystyle{unsrt}
\bibliography{refer/alias,refer/main}
}


\newpage
\appendix
\section{Network Architecture}
\label{sec:app_arch}
 
This appendix section expands on the architecture introduced in \cref{sec:formulation}.
We detail the camera parameterization that \cref{sec:formulation} packages as $\mathbf{C}$, the backbone that realize $G_\theta$, the decoder heads that emit output of $\mathcal{Y}$, and the optional conditioning pathways for known camera parameters and user prompts.
 
\subsection{Camera Parameterization}
\label{sec:app_camera}

The per-frame camera variable $\mathbf{C}_i = (\mathbf{K}_i, \mathbf{R}_i, \mathbf{t}_i)$ packages intrinsics, rotation, and translation.
The network emits $\mathbf{C}_i$ independently per frame as a 13-dimensional vector.
Six dimensions parameterize rotation $\mathbf{R}_i$ via the continuous 6D representation~\cite{zhou2019continuity}, three dimensions encode translation $\mathbf{t}_i$, and four dimensions encode normalized intrinsics $(\nicefrac{f_x}{W},\, \nicefrac{f_y}{H},\, \nicefrac{c_x}{W},\, \nicefrac{c_y}{H})$ for $\mathbf{K}_i$.

The 6D representation avoids the antipodal ambiguity of quaternions and the discontinuities of Euler angles.
A Gram-Schmidt step recovers a valid rotation matrix from the regressed six values.
Normalizing $f_x, c_x$ by image width $W$ and $f_y, c_y$ by image height $H$ keeps the parameterization scale-invariant across resolutions, so training data of different image sizes mixes without rescaling.

We anchor the world coordinate frame to the first frame, fixing $\mathbf{R}_0 = \mathbf{I}$ and $\mathbf{t}_0 = \mathbf{0}$.
The remaining $(\mathbf{R}_i, \mathbf{t}_i)$ then describe relative camera motion with respect to the first frame.
This convention defines a consistent world coordinate frame even in the monocular setting where global scale and orientation are otherwise unobservable.

A single physical camera produces every frame in the clip, yet we do not hard-tie $\mathbf{K}_i$ across frames.
The network outputs $\mathbf{K}_i$ independently per frame, and the camera-consistency term in \cref{sec:app_anchor} softly drives them to agree.
The soft formulation accommodates digital zoom or cropping, where intrinsics drift over time.
When known camera parameters are available, the conditioning pathway in \cref{sec:app_cond} consumes them directly in place of these outputs.

\subsection{Backbone and Trunk}
\label{sec:app_backbone}

We adopt DINOv3 ViT-L/16~\cite{simeoni2025dinov3} pretrained on LVD-1689M as our visual backbone and keep it frozen throughout training.
The frozen backbone preserves general-purpose visual priors while concentrating trainable capacity on the trunk and downstream heads.
Each input frame is resized to $512 \times 512$ and tokenized into a $32 \times 32$ patch grid, producing 1024 patch tokens per frame at hidden dimension $D = 1024$.
Backbone forward passes run in bf16 precision.

A lightweight projector sits between the frozen backbone and the trunk.
It applies 2 spatial-adaptation layers to patch tokens alone, performing intra-frame attention to map frozen features into the trunk input space.
This module adapts the general-purpose representation toward downstream geometric tasks without unfreezing the backbone.

The trunk consists of 16 joint query-patch attention layers (\cref{fig:architecture}).
Pose queries and camera queries enter the attention alongside patch tokens at every layer, so queries interact with dense features throughout the stack rather than at a single final readout.
Each trunk layer factorizes attention into a frame attention sub-module and a temporal attention sub-module.
Frame attention performs intra-frame spatial attention to capture per-frame structure.
Temporal attention aggregates across frames, applying 1D temporal RoPE on Q and K to encode frame indices and capture temporal dynamics.
All trunk layers use pre-norm LayerNorm with 16 attention heads, and we run SDPA kernels with FlashAttention-2 on H100 GPUs.

For a clip of length $T$, the trunk processes $T \cdot (1024 + J + 1) + N_{\text{cond}}$ tokens in total, where $J = 24$ joint queries follow the SMPL skeleton convention and the camera query contributes one token per frame.
The optional conditioning tokens $N_{\text{cond}}$ are detailed in \cref{sec:app_cond}.

\subsection{Decoder Heads}
\label{sec:app_heads}

The dense heads emit depth $\mathbf{D}$, flow $\mathbf{F}$, and mask $\mathbf{M}$ as pixel-level outputs.
We instantiate three independent DPT-style~\cite{ranftl2021vision} multi-scale fusion modules, one per modality.
Each head taps four trunk layers at $\{4, 8, 12, 16\}$ for uniform $\nicefrac{1}{4}$ spacing, fuses them at 256-d feature width, and bilinearly upsamples from the $32 \times 32$ patch grid to the original $512 \times 512$ resolution.
Output activations differ by physical range: softplus on $\mathbf{D}$ enforces positive depth, identity on $\mathbf{F}$ outputs a signed 3D vector field, and sigmoid on $\mathbf{M}$ yields soft foreground probability.

The sparse heads emit pose $\mathbf{P}$ and camera $\mathbf{C}$ from the final-layer query states.
Each pose query corresponds to one of $J = 24$ skeletal joints, and a 3-layer MLP maps its final state to $\mathbb{R}^3$ joint coordinates.
The per-frame camera query passes through a 3-layer MLP that emits the 13-dimensional vector parsed under \cref{sec:app_camera}'s convention.
Joint angles required by the minimum-effort-path constraint are derived from these joint coordinates via inverse kinematics, detailed in \cref{sec:app_losses}.

The three dense heads maintain independent fusion features rather than sharing a single tower.
This isolates per-modality gradients within each head, preventing high-frequency texture cues that benefit one modality from polluting the geometric structure required by another.
 
\subsection{Optional Conditioning}
\label{sec:app_cond}

When ground-truth camera parameters $(\mathbf{K}, \mathbf{R}, \mathbf{t})$ are available, the system consumes them in place of the camera head's output.
Known parameters are packaged into the 13-d format of \cref{sec:app_camera}, encoded by a 3-layer MLP into a single token, and added to the corresponding camera query state for that frame.
The additive fusion anchors the model to accurate camera geometry when prior knowledge exists and falls back to self-prediction otherwise.

A user can optionally specify the subject of interest through 2D points and a bounding box.
We follow the SAM~\cite{kirillov2023segment} prompt encoder design: up to 5 points and one 2-corner box per frame, with positions encoded via fixed random Fourier features and types (foreground point, background point, box corner) encoded via learnable label embeddings.
All prompt tokens enter the trunk attention as additional tokens, interacting with patch, pose, and camera queries on equal footing.

We randomly drop conditioning inputs during training following the augmentation recipe of SAM~\cite{kirillov2023segment}.
The camera condition is replaced by a learnable null token with probability 0.5.
Each prompt point slot is dropped independently with probability 0.3, and the box is dropped with probability 0.5.
The model thus encounters a wide range of conditioning combinations during training, producing valid outputs at inference whether the user supplies full conditioning, partial prompts, or no conditioning at all.
 
\section{Self-supervised Learning}
\label{sec:app_bio}

This appendix section expands on the four constraints and auxiliary anchoring terms introduced in \cref{sec:loss}.
We give the implementation forms of tolerances, anatomical constants, support-polygon construction, and inverse/forward kinematics.
We then specify the auxiliary distillation and camera-consistency losses, and close with a coupling footprint that summarizes which output modalities each prior touches.
 
\subsection{Loss Implementation Details}
\label{sec:app_losses}

This subsection specifies the implementation forms of the four priors in \cref{sec:loss}, covering tolerance values, anatomical constants, support-polygon construction, and the inverse and forward kinematics conventions used by the minimum-effort-path constraint.

\paragraph{Silhouette Edge Alignment}
The signed distance field $\mathrm{SDF}_{\partial \mathbf{M}}$ in \cref{eq:loss-silh} grows unboundedly away from the boundary and would otherwise dominate the gradient sum at distant pixels.
We saturate its absolute value at $\tau = 32$ pixels via $|\widetilde{\mathrm{SDF}}_{\partial \mathbf{M}}| = \min(|\mathrm{SDF}_{\partial \mathbf{M}}|, \tau)$, concentrating the constraint on the transition band near the boundary rather than the far field.
The spatial gradients $\nabla \mathbf{D}$ and $\nabla \mathbf{F}$ are computed by a Sobel $3 \times 3$ operator over bf16 features.

\paragraph{Skeletal-Surface Coupling}
The tolerance follows a piecewise-linear form $\rho(\mathbf{u}) = \rho_{\min} + \alpha \cdot d_{\text{bone}}(\mathbf{u})$, where $d_{\text{bone}}(\mathbf{u})$ is the 3D distance from the unprojected surface point to its nearest bone segment.
We set $\rho_{\min} = 1$ cm as the on-skin floor and $\alpha = 0.5$ as the dimensionless growth rate, so on-skin pixels stay tight to the skeleton while loose garments such as a skirt receive a wider margin.
Bone assignment $b^* = (j_1^*, j_2^*)$ and barycentric position $\alpha^* \in [0, 1]$ are obtained via $\arg\min$ over bone segments, which is non-differentiable.
We carry stop-gradient through the selection, allowing gradients only through the flow residual and the tolerance term.

\paragraph{Center-of-Mass Support}
Anatomical mass distribution constants $\{w_j\}_{j=1}^{24}$ follow the segmental inertia table of De Leva~\cite{deleva1996adjustments}, normalized so that $\sum_j w_j = 1$.
The contact weight $c_j = \exp(-d_j^2 / \sigma^2)$ uses end-effector height $d_j$ above a fitted ground plane with $\sigma = 5$ cm.
The ground plane is recovered from $\mathbf{D}$ by RANSAC over the lowest non-foreground point cloud, restricted to lie below the foreground bounding box.
Support polygon $\mathcal{S}(\mathbf{P}, \mathbf{D})$ is the convex hull of the end-effector ground projections $\{\mathbf{P}_{j,xy}\}$ weighted by $c_j$, with the SDF evaluated on the weighted polygon boundary $\partial \mathcal{S}$.
The mask $\mathbf{M}$ further restricts candidate contact joints to the foreground, preventing spurious contacts from background limbs.

\paragraph{Minimum Effort Path}
Inverse kinematics traverse the kinematic chain from the pelvis root to the leaves, encoding each parent-child rotation in local axis-angle form with bone roll fixed to zero to resolve the rotation ambiguity along the bone axis.
Forward kinematics apply the standard parent-to-child rotation chain.
The temporal window $W$ is sampled per training iteration from the clip index set $\mathcal{T}$ with random length $|W| \in [W_{\min}, W_{\max}]$, where $W_{\min} = 8$ frames and $W_{\max} = 32$ frames.
The trajectory tolerance is $\rho_{\text{eff}} = 5$ cm, beyond which deviations from the minimum-jerk reference incur penalty.
 
\subsection{Anchoring and Camera Consistency}
\label{sec:app_anchor}

This subsection specifies the distillation anchor and the camera consistency term mentioned at the end of \cref{sec:loss}.

\paragraph{Distillation Anchor}
The four physical priors in \cref{sec:loss} are scale-invariant by construction and cannot pin down the global metric scale of the network's outputs.
We anchor depth $\mathbf{D}$ and pose $\mathbf{P}$ against two off-the-shelf teachers, Depth Pro~\cite{bochkovskii2024depthpro} for $\mathbf{D}_{\text{teach}}$ and SAM 3D Body~\cite{yang2026sam3dbody} for $\mathbf{P}_{\text{teach}}$, through a margin-based hinge:
\begin{equation}
\mathcal{C}_{\text{dist}}(\mathbf{D}, \mathbf{P}) = \frac{1}{|\Omega|} \sum_{\mathbf{u}} \big[\, |\mathbf{D}(\mathbf{u}) - \mathbf{D}_{\text{teach}}(\mathbf{u})| - \rho_{\mathbf{D}} \,\big]_{+} + \frac{1}{J} \sum_{j=1}^{J} \big[\, \|\mathbf{P}_{j} - \mathbf{P}_{\text{teach}, j}\|_{2} - \rho_{\mathbf{P}} \,\big]_{+},
\label{eq:dist}
\end{equation}
with $\rho_{\mathbf{D}} = 10$ cm and $\rho_{\mathbf{P}} = 3$ cm.
Teachers are noisy, and a strict regression target would collapse \paper into a teacher replica.
The margin instead activates the anchor only when \paper drifts beyond a tolerance band, leaving the physical priors free to optimize within that band.

\paragraph{Camera Consistency}
A single physical camera produces every frame in the clip, yet \cref{sec:app_camera} predicts $\mathbf{K}_i$ independently per frame to preserve flexibility for digital zoom or cropping.
A consistency term softly drives the per-frame intrinsics to agree:
\begin{equation}
\mathcal{C}_{\text{cam}}(\mathbf{C}) = \frac{1}{|\mathcal{T}|^{2}} \sum_{i, j \in \mathcal{T}} \|\mathbf{K}_{i} - \mathbf{K}_{j}\|_{2}^{2},
\label{eq:cam}
\end{equation}
where $\mathbf{K}_i$ takes the 4-d normalized form of \cref{sec:app_camera} so that the squared norm is scale-aligned across resolutions.
Per-frame rotations and translations $(\mathbf{R}_i, \mathbf{t}_i)$ are unconstrained, since they describe inter-frame camera motion by design.

\section{Implementation}
\label{sec:app_impl}

\paragraph{Training Setup}
We train \paper on 6 NVIDIA H100 80GB GPUs with mixed precision (bf16 forward, fp32 master weights) for $4 \times 10^{5}$ iterations.
The optimizer is AdamW with base learning rate $4 \times 10^{-5}$, weight decay $0.05$, and a cosine schedule with linear warmup over the first 5K iterations.
The effective batch size is 12 clips, distributed as 2 clips per GPU across 6 GPUs, with each clip containing $T = 16$ frames.
Gradient clipping is applied with threshold 1.0.

\paragraph{Data Pipeline}
Pretraining draws from all seven Human3.6M subjects.
Each mini-batch randomly samples 16 consecutive frames at a stride drawn uniformly from a small range, exposing the model to a spread of inter-frame intervals rather than a single fixed frame rate.
This dynamic temporal sampling lets \paper generalize across datasets with different native frame rates without retraining.
Augmentations include foreground-centered random crops, horizontal flipping, and color jitter.
For in-domain fine-tuning on Fit3D and 3DPW, we follow the splits described in \cref{sec:experiments} and reuse the pretraining hyperparameters with a shorter $1 \times 10^{5}$ iteration schedule.

\paragraph{Inference}
At test time, \paper accepts video clips of arbitrary length.
Clips at or below the training length pass through the network in a single forward call.
Longer clips are processed via a sliding window of 16 frames with 4-frame overlap between adjacent windows, and predictions in the overlap region are averaged across windows.
All output heads are produced in a single forward pass, with no test-time optimization or runtime calibration.

\section{Additional Qualitative Results}
\label{sec:additional_qualitative}

We provide multi-view renderings of \paper alongside Zero MSF~\cite{liang2025zeroshot} across three settings of increasing distribution shift: an in-the-wild ballet performance (\cref{sec:app_teaser}), in-domain Fit3D sequences (\cref{sec:app_fit3d}), and out-of-domain DynAct4D sequences with heavy garment deformation (\cref{sec:app_dynact}).

\subsection{Multi-view Renderings of the Ballet Reconstruction}
\label{sec:app_teaser}

\begin{figure}[t]
    \centering
    \includegraphics[width=\linewidth]{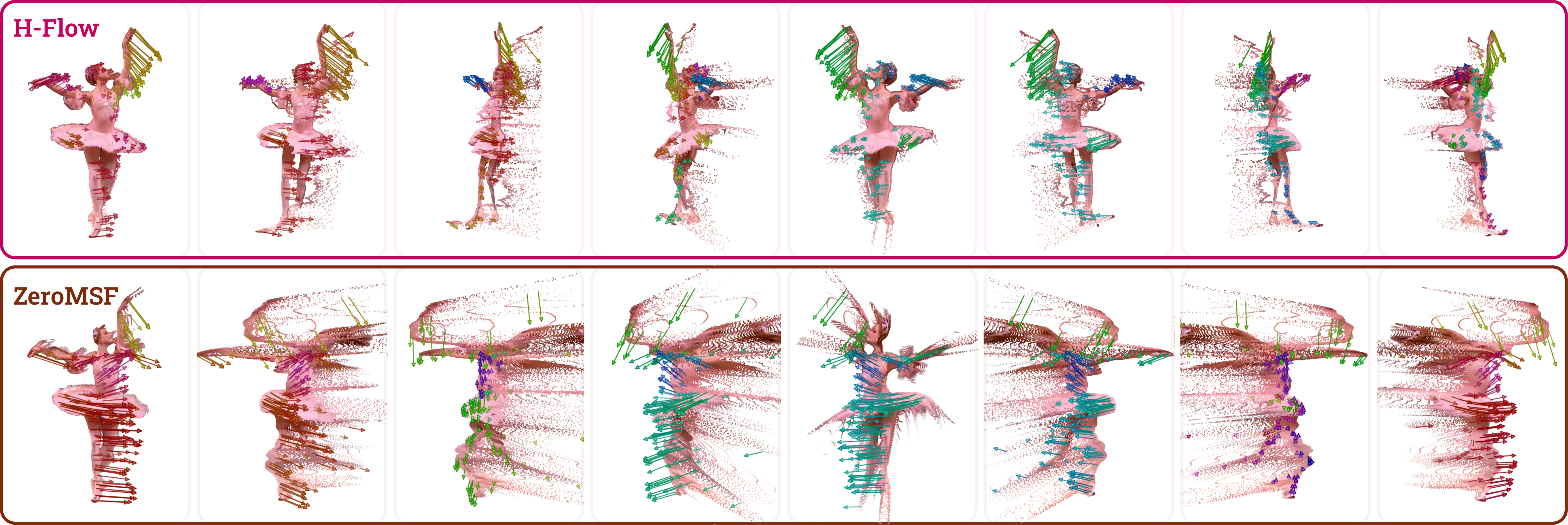}
    \caption{\macaption{Multi-view renderings of the ballet reconstruction.}
        \micaption{The 3D point cloud from \cref{fig:teaser}, rotated around the dancer at additional azimuths.
        Top: Zero MSF~\cite{liang2025zeroshot}.
        Bottom: \paper.
        Elevation and field of view are held fixed across all renderings.}}
    \label{fig:app_teaser}
\end{figure}

\Cref{fig:teaser} renders each method's predicted depth as a textured 3D point cloud from a single frontal viewpoint.
This view obscures whether the reconstructed geometry is genuinely volumetric or merely a frontal-facing depth sheet.
\Cref{fig:app_teaser} rotates the same scene around the dancer at additional azimuths.
Zero MSF collapses off-axis into a near-planar surface, while \paper preserves coherent volumetric structure across all viewpoints, with the tutu and extended limbs remaining distinguishable from the background.
The corresponding orbit animations are provided in the supplementary material.

\subsection{Multi-view Renderings on Fit3D (In-Domain)}
\label{sec:app_fit3d}

\begin{figure}[t]
    \centering
    \includegraphics[width=\linewidth]{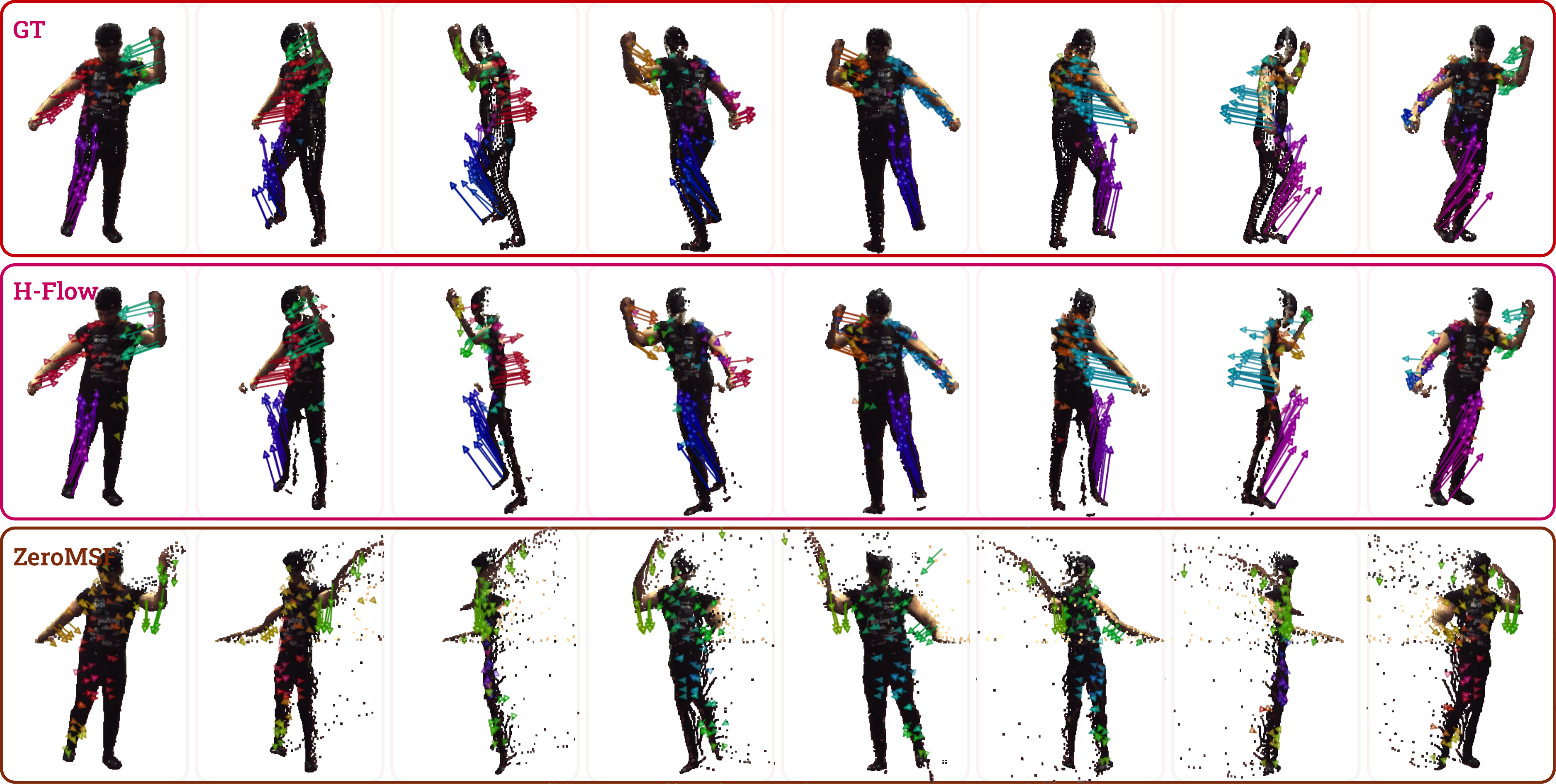}
    \caption{\macaption{Multi-view flow comparison on Fit3D.}
        \micaption{Each column renders the same predicted flow from a different azimuth.}}
    \label{fig:app_fit3d}
\end{figure}

\cref{fig:app_fit3d} shows multi-view flow predictions on Fit3D.
\paper produces flow vectors that align with the ground truth in both direction and magnitude across all viewpoints, on the torso and the limbs alike.
ZeroMSF returns fragmented point clouds and sparse, misdirected vectors, leaving the subject's articulation largely unregistered.

\subsection{Multi-view Renderings on DynAct4D (Out-of-Domain)}
\label{sec:app_dynact}

\begin{figure}[t]
    \centering
    \includegraphics[width=\linewidth]{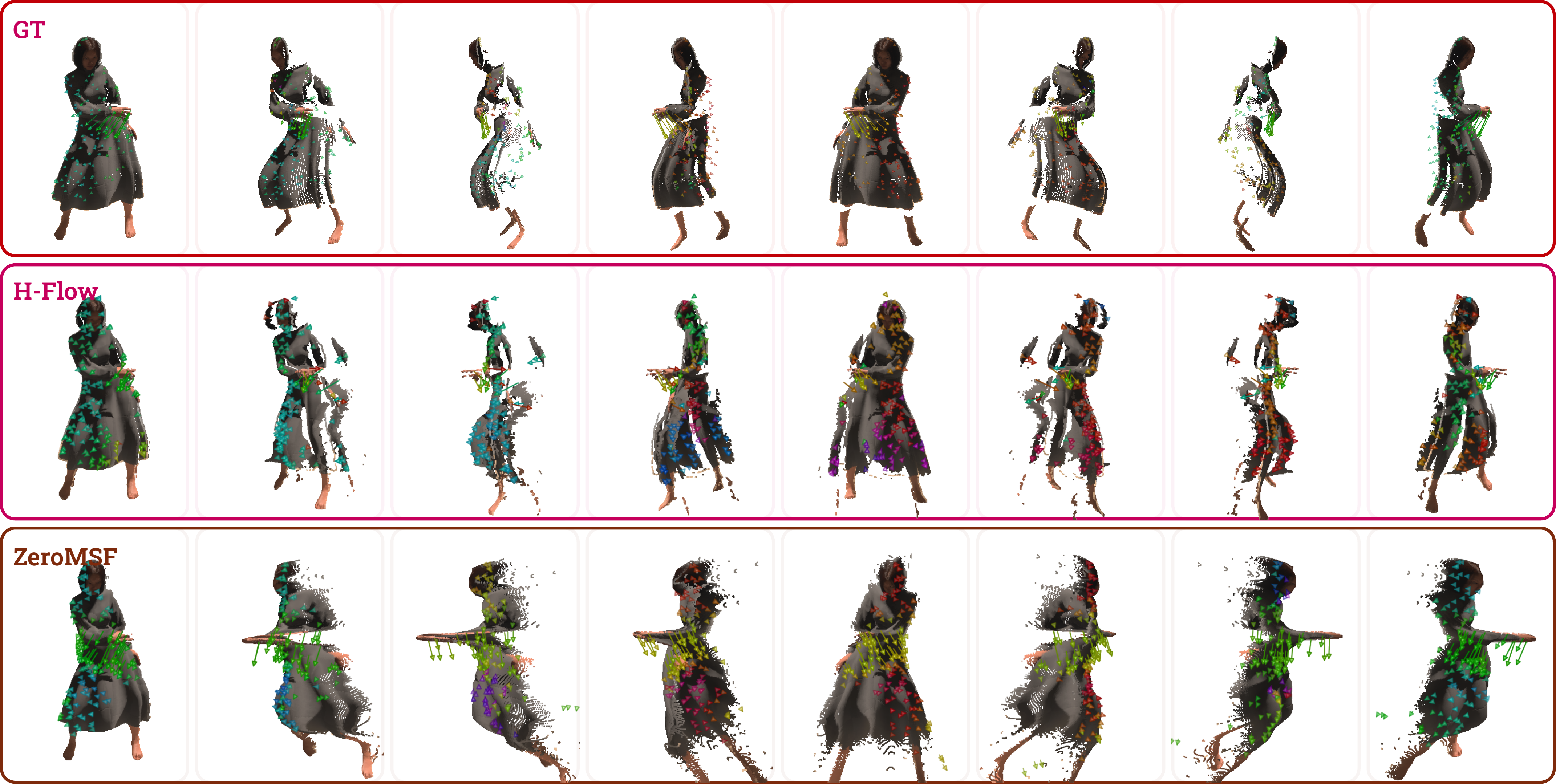}
    \caption{\macaption{Multi-view flow comparison on DynAct4D.}
        \micaption{A sequence with significant garment deformation, viewed from multiple azimuths.}}
    \label{fig:app_dynact}
\end{figure}

\cref{fig:app_dynact} extends the comparison to DynAct4D, where a long dress introduces non-rigid deformation that lies outside the training distribution.
\paper recovers a coherent body silhouette across viewpoints and produces structurally consistent flow on both the torso and the skirt, capturing the garment's swirl alongside the underlying body motion.
Zero MSF fragments the point cloud, concentrates sparse flow on the upper body, and leaves the skirt's deformation largely unmodeled.

\end{document}